\documentclass[graybox]{svmult}

\usepackage{mathptmx}       
\usepackage{helvet}         
\usepackage{courier}        
\usepackage{type1cm}        
\usepackage{makeidx}         
\usepackage{graphicx}        
\usepackage{multicol}        
\usepackage[bottom]{footmisc}

\usepackage[utf8]{inputenc}
\usepackage{todonotes}
\usepackage{multirow}
\usepackage{tablefootnote}
\usepackage{natbib} 
\usepackage[hyphens]{url} 
\definecolor{blue}{RGB}{0,0,0} 

\makeindex             

\begin{document}

\title*{What Level of Quality can Neural Machine Translation Attain on Literary Text?}
\author{Antonio Toral \and Andy Way}
\institute{Antonio Toral \at Center for Language and Cognition, University of Groningen, The Netherlands, \email{a.toral.ruiz@rug.nl} \and
Andy Way \at ADAPT Centre, School of Computing, Dublin City University, Ireland, \email{andy.way@adaptcentre.ie}}
%
%
\maketitle


\abstract{
Given the rise of a new approach to MT, Neural MT (NMT), and its promising performance on different text types, we assess the translation quality it can attain on what is perceived to be the greatest challenge for MT: literary text.
Specifically, we target novels, arguably the most popular type of literary text.
We build a literary-adapted NMT system for the English-to-Catalan translation direction and evaluate it against a system pertaining to the previous dominant paradigm in MT: statistical phrase-based MT (PBSMT).
To this end, for the first time we train MT systems, both NMT and PBSMT, on large amounts of literary text (over 100 million words)  and evaluate them on a set of twelve widely known novels spanning from the the 1920s to the present day. 
According to the BLEU automatic evaluation metric, NMT is significantly better than PBSMT ($p<0.01$) on all the novels considered.
Overall, NMT results in a 11\% relative improvement (3 points absolute) over PBSMT. 
A complementary human evaluation on three of the books shows that between 17\% and 34\% of the translations, depending on the book, produced by NMT (versus 8\% and 20\% with PBSMT) are perceived by native speakers of the target language to be of equivalent quality to  translations produced by a professional human translator.
}

\section{Introduction}\label{s:introduction}

Literary text is considered to be the 
greatest challenge
for machine translation (MT).
According to perceived wisdom, despite the tremendous progress in the field of statistical MT over the past two decades, 
there is no prospect of machines being useful at (assisting with) the translation of this type of content.

However, we believe that the recent emergence of two unrelated technologies opens a window of opportunity to explore this topic:

\begin{enumerate}
\item The electronic book: the market share of e-books is continuously growing,\footnote{For example, in the US the market share of e-books surpassed that of printed books for fiction in 2014, \url{http://www.ingenta.com/blog-article/adding-up-the-invisible-ebook-market-analysis-of-author-earnings-january-2015-2/}}
as a result of which there is a wide availability of
books in digital format, including original novels and their translations. Because the main resource required to train statistical MT
systems is bilingual parallel text, we are now able to build MT systems tailored to novels.
This should result in better performance, as it has been shown in MT research again and again that for a statistical MT engine to perform optimally it should be trained on similar data to the data it is applied to, e.g.~\cite{Pecina2014}.

\item Neural MT: NMT is a new approach to statistical MT, which,  while having been introduced only very recently,\footnote{Working models of NMT have only recently been introduced, but from a theoretical perspective, very similar models can be traced back two decades \citep{Forcada1997}.} has already shown great potential, as there is evidence that it can attain better translation quality than the dominant approach to date, namely phrase-based statistical MT (PBSMT).
This has been shown for a number of language pairs and domains, including transcribed speeches~\citep{Luong-Manning:iwslt15}, newswire~\citep{sanchezcartagena-toral:2016:WMT} and United Nations documents~\citep{junczys30}.
Beyond its generally positive performance, NMT is of particular interest for literary texts due to the following two findings:
	\begin{itemize}
	\item Its performance seems to be especially promising for lexically-rich texts~\citep{bentivogli-EtAl:2016:EMNLP2016}, which is the case of literary texts.
	\item There are claims that NMT ``can, rather than do a literal translation, find the cultural equivalent in another language''.\footnote{\url{http://events.technologyreview.com/video/watch/alan-packer-understanding-language/}} 
	\end{itemize}
\end{enumerate}

With respect to the last point, literal (but not word-for-word) translations are deemed acceptable for domains for which PBSMT is already widely used in industry, such as technical documentation,  as the aim of the translation process here is purely to carry over the meaning of the source sentence to the target language, without necessarily reflecting any stylistic niceties of the target language.
In contrast, literal translations are not at all suitable for literary texts because 
the expectations of the reader are considerably higher; it is not sufficient for the translation to merely preserve the meaning, as it should also preserve the reading experience of the original text.

In this chapter we aim to assess the performance that can be offered by state-of-the-art MT for literary texts.
To this end we train PBSMT and NMT systems for the first time on large amounts of literary texts (over 100 million words) and evaluate them on a set of twelve widely known novels that span from the 1920s to the beginning of the 21$^{st}$ century.

The rest of the chapter is organised as follows.
In the following section, we provide an overview of the research carried out in the field of MT targeting literary texts.
Next, we outline our experimental setup (Section \ref{s:experimental_setup}) and provide technical details of the PBSMT and NMT systems built (Section \ref{s:mt_systems}).
Subsequently we evaluate and analyse the translations produced by both MT systems (Section \ref{s:evaluation}).
Finally, we conclude and outline lines of future work in Section \ref{s:conclusions}.

\section{State-of-the-Art in MT of Literary Text}\label{s:state_of_the_art}

There has been recent interest in the Computational Linguistics community regarding the processing of
literary text. The best example is the establishment of an annual workshop (Computational Linguistics for Literature) in 2012.
A popular strand of research concerns the automatic identification of text snippets that convey figurative devices, such as metaphor~\citep{Shutova:2013:SMP:2483810.2483813}, idioms~\citep{Li:2010:UGM:1857999.1858038}, humour and irony~\citep{DBLP:journals/pdln/Reyes13}, applied to monolingual text.
Conversely, there has been a rather limited amount of work on applying MT to literary texts, as we now survey.

\cite{Genzel:2010:PSM:1870658.1870674} constrained SMT systems for poetry to produce French-to-English translations that obey length, meter, and rhyming rules. Form is preserved at the price of producing considerably lower quality translations; the score according to the BLEU automatic evaluation metric~\citep{Papineni:2002:BMA:1073083.1073135} (see Castilho et al. and Way in this volume for more details of this metric) decreases by around 50\%, although it should be noted that their evaluation was not on poetry but on news. 

\cite{greene2010automatic} translated poetry, choosing target output realisations that conform to the
desired rhythmic patterns.
Specifically, they translated Dante's {\it Divine Comedy} from Italian sonnets into English iambic pentameter.
Instead of constraining the SMT system, as done by \cite{Genzel:2010:PSM:1870658.1870674}, they passed its output lattice through a device that maps words to sequences of stressed and unstressed syllables.
These sequences were finally filtered with an iambic pentameter acceptor. 


\cite{W12-2503} examined the role of referential cohesion in translation and found that literary texts have more dense reference chains. 
They concluded that incorporating discourse features beyond the level of the sentence~\citep{hardmeier2014discourse} is an important research focus for applying MT to literary texts.

\cite{jones2013faithful} used general-domain MT systems to translate samples of French literature (prose and poetry) into English. They then used qualitative analysis grounded in translation theory on the MT output to assess the potential of MT in literary translation and to address what makes literary translation particularly difficult.

\cite{besacier:hal-01003944} used MT followed by post-editing (by non-professional translators) to translate a short story from English into French. Such a workflow was deemed a useful low-cost alternative for translating literary works, albeit at the expense of disimproved translation quality.

Our recent work \citep{toral-way:2015:CLfL} contributed to the state-of-the-art in two dimensions.
First, we conducted a comparative analysis on the translatability of literary text according to narrowness of the domain and freedom of translation, which is more general and complementary to the analysis by \cite{W12-2503}.
Second, related to \cite{besacier:hal-01003944}, we evaluated MT for literary text. There were two differences though; first, \cite{besacier:hal-01003944} translated a short story, while we translated a novel; second, their MT systems were evaluated against a post-edited reference produced by non-professional translators, while we evaluated our MT systems against the translation produced by a professional translator.




\textcolor{blue}{
This work builds upon our previous study~\citep{toral-way:2015:CLfL}, the following being the main differences between the two: we now train a literary-adapted MT system under the NMT paradigm (while previously we used PBSMT), the translation direction considered is more challenging as the languages are more distant (English-to-Catalan versus Spanish-to-Catalan), we conduct a considerably broader evaluation (12 books now versus 1 in the previous work) and analyse the results with respect to a set of textual features of each novel.
}

\section{Experimental Setup}\label{s:experimental_setup}

This section covers the experimental settings.
We explain the motivation for the language pair chosen for this chapter (Section \ref{s:language_pair}), describe the data sets used in our experiments (Section \ref{s:data_sets}) and finally the tools that were utilised (Section \ref{s:tools}).

\subsection{Language Pair}\label{s:language_pair}
In general, it is widely accepted that the quality attainable by MT correlates with the level of relatedness between the pair of languages involved.
This is because translations between related languages should be more literal, and complex phenomena (such as metaphorical expressions) might simply transfer rather straightforwardly to the target language, while they are more likely to require complex translations between unrelated languages.

In our previous work~\citep{ts2015,toral-way:2015:CLfL}, we considered a closely-related language pair (Spanish-to-Catalan), where both languages belong to the same family (Romance).
We built a literary-adapted PBSMT system and used it to translate a novel from an internationally renowned author, Ruiz Zafón.
We concluded that our system could be useful to assist with the translation of this kind of text due to the following two findings.
\begin{enumerate}
\item For a random subset of sentences from the novel, we asked native speakers to rank the translations coming from the MT system against those from a professional translator (i.e. taken from the published novel in the target language), although they did not know which were which. For over 60\% of the sentences, native speakers found both translations to be of the same quality~\citep{toral-way:2015:CLfL}.

\item The previous evaluation was carried out at the sentence level, so it might be argued that this is somewhat limited as it does not take context beyond the sentence into account. Therefore we subsequently analysed 3 representive passages (up to 10 consecutive sentences): one of average MT quality (i.e. the quality of this passage is similar to the quality obtained by MT on the whole novel, as measured with BLEU), another of high quality (i.e. its BLEU score is similar to the average BLEU score of the 20\% highest-scoring passages) and finally, one of low quality (i.e. its BLEU score is similar to the average BLEU score of the 20\% lowest-scoring passages). For the passages of high and average quality, we showed that the MT output requires only a few character edits to match the professional translation~\citep{ts2015}.
\end{enumerate}

Encouraged by the positive results obtained on a closely-related language pair, we have now decided to explore the potential for a less-related pair - correspondingly a more challenging task.
The language pair in this study is English-to-Catalan, where the two languages involved belong to different families (Germanic and Romance, respectively).

We choose Catalan as target language as an example of a mid-size European language.\footnote{With this term we refer to European languages with around 5 to 10 million speakers, as is the case of many other languages in Europe, such as Danish, Serbian, Czech, etc.}
These are languages into which a significant number of novels have been translated - we have easily identified over 200 English e-books available in Catalan. Nonetheless, this number is very low compared to the amount of books translated into `major' European languages (such as German, French, Italian, or Spanish).
Concerning mid-size European languages, because there is a reasonable amount of data available to train literary-adapted MT systems and also room to have more novels translated if the output translations produced by MT are deemed useful to assist translators, we believe this is a sensible choice of target language type for this line of research.

\subsection{Data sets}\label{s:data_sets}

\subsubsection{Training and Development Data}

We use parallel and monolingual in-domain data for training.
The parallel data comprises 133 parallel novels (over 1 million sentence pairs),
while the monolingual data consists of around 1,000 books written in Catalan (over 5 million sentences) and around 1,600 books in English\footnote{While our experiments are for the English-to-Catalan language pair, we also use English monolingual data to generate synthetic data for our NMT system (see Section \ref{s:nmt}).} (over 13 million sentences).
In addition, we use out-of-domain datasets, namely OpenSubtitles\footnote{\url{http://opus.lingfil.uu.se/OpenSubtitles.php}} as parallel data (around 400,000 sentence pairs) and monolingual Catalan data (around 16 million sentences) crawled from the web~\citep{LJUBEI14.841}.
The development data consists of 2,000 sentence pairs randomly selected from the in-domain parallel training data and removed from the latter data set.
Quantitative details of the training and development data sets are shown in Table \ref{t:datasets}.

\begin{table}[h]
\begin{center}
\begin{tabular}{lrrr}
\hline
\multirow{2}{*}{\bf Dataset} & \multirow{2}{*}{\bf \# sentences} & \multicolumn{2}{c}{\bf\# tokens}\\
					&			&\bf English	& \bf Catalan\\
\hline
Training parallel (in-domain)		&1,086,623		&16,876,830	&18,302,284	\\
Training parallel (OpenSubs)		&402,775		&3,577,109		&3,381,241\\
\hline
Training monolingual (in-domain)					&5,306,055		&-				&100,426,922\\
Training monolingual (in-domain)	&13,841,542	&210,337,379	&-\\
Training monolingual (web)						&16,516,799	&-				&486,961,317\\
\hline
Development						& 2,000			&34,562		&38,114\\
\hline
\end{tabular}
\end{center}
\caption{Number of sentences and tokens (source and target sides) in the training and development data sets.}
\label{t:datasets}
\end{table}

\subsubsection{Test Data}

We test our systems on 12 English novels and their professional translations into Catalan.
In so doing we aim to build up a representative sample of literary fiction, encompassing novels from different periods (from the 1920s to the present day) and genres and targeted at different audiences.
Details are provided in Table \ref{t:testset}.
For each novel, aside from the number of sentences and tokens (i.e. words) that it contains, we show also the portion of the source book (percentage of sentences) that was evaluated.\footnote{In order to build the test sets we sentence-align the source and target versions of the books. We keep the subset of sentence pairs whose alignment score is above a certain threshold. See Section \ref{s:preprocessing} for further details.}

\begin{table}[h]
\begin{center}
\begin{tabular}{lrrrr}
\hline
\multirow{2}{*}{\bf Author, book and year} & \multirow{2}{*}{\bf \% sentences} & \multirow{2}{*}{\bf \# sentences} & \multicolumn{2}{c}{\bf\# tokens}\\
						      &&	   									&\bf English	& \bf Catalan\\
\hline
Auster's {\it Sunset Park}	(2010)				&75.43\%	&2,167	&70,285	&73,541\\
Collins' {\it Hunger Games \#3} (2010)				&73.36\%	&7,287	&103,306	&112,255	\\
Golding's {\it Lord of the Flies} (1954)				&82.93\%	&5,195	&64,634	&69,807\\
Hemingway's {\it The Old Man and the Sea} (1952)&76.01\%	&1,461	&24,233	&25,765\\
Highsmith's {\it Ripley Under Water} (1991) 		& 65.86\%	&5,981	&84,339	&94,565\\
Hosseini's {\it A Thousand Splendid Suns} (2007) 	& 67.54\%	&6,619	&97,728	&105,989\\
Joyce's {\it Ulysses} (1922)							& 46.65\%	&11,182 &136,250	&159,460\\ 
Kerouac's {\it On the Road} (1957)					& 76.35\%	&5,944	&106,409	&111,562\\
Orwell's {\it 1984} (1949)							& 68.23\%	&4,852	&84,062	&90,545	\\
Rowling's {\it Harry Potter \#7} (2007)				& 69.61\%	&10,958&186,624	&209,524\\
Salinger's {\it The Catcher in the Rye} (1951) 		& 76.57\%	&5,591	&77,717	&77,371\\
Tolkien's {\it The Lord of the Rings \#3} (1955) 			& 66.60\%	&6,209	&114,847	&129,671\\
\hline
\end{tabular}
\end{center}
\caption{Percentage of sentences used from the original data set and number of sentences and tokens in the novels that make up the test set.}
\label{t:testset}
\end{table}

Whilst obviously none of the novels in the test set is included in the training data, the latter dataset may contain other novels from writers represented in the test set.
For example, the test set contains the 7$^{th}$ book in the {\it Harry Potter} series from J. K. Rowling and the training set contains the previous 6 books of that series.
Table \ref{t:prev_books} shows, for each writer represented in the test set, how many books appear in the training set from this writer, and how many sentence pairs and tokens (source side) these books amount to.

\begin{table}[h]
\begin{center}
\begin{tabular}{lrrr}
\hline
\bf Author	&\bf \# Books	&\bf \# sentence pairs	&\bf\# tokens (English)\\
\hline
Auster		&2	 	&6,831			&145,195\\
Collins		&2	 	&15,315			&216,658\\
Golding		&0		&0			&0\\
Hemingway	&0		&0			&0\\
Highsmith	&4		&27,024			&382,565\\
Hosseini	&1		&7,672			&105,040\\
Joyce		&2		&8,762			&146,525\\
Kerouac		&0		&0			&0\\
Orwell		&2		&4,068			&88,372\\
Rowling		&6		&50,000			&836,942\\
Salinger	&4		&8,350			&141,389\\
Tolkien		&3		&23,713			&397,328\\
\hline
\end{tabular}
\end{center}
\caption{Number of books in the training set, together with their overall number of sentence pairs and source-side tokens for each writer that is also represented in the test set.}
\label{t:prev_books}
\end{table}


\subsection{Tools}\label{s:tools}
We have leveraged state-of-the-art techniques in the field through the pervasive use of open-source tools throughout the different stages of our experimentation, namely preprocessing, MT experimentation and evaluation, as detailed in the remainder of this section.

\subsubsection{Preprocessing}\label{s:preprocessing}

The datasets (see Section \ref{s:data_sets}) are preprocessed in order to make them suitable for MT.
In-domain data is extracted from e-books and converted to plain text with Calibre support tools,\footnote{\url{https://calibre-ebook.com/}} then sentence-split with NLTK~\citep{bird2006nltk} and Freeling~\citep{padro2012freeling} for English and Catalan, respectively,
subsequently tokenised with Moses' scripts and Freeling, for English and Catalan, respectively,
and finally sentence-aligned with Hunalign~\citep{hunalign}.
Sentence alignment is carried out on lowercased text, in order to reduce data sparsity, with the assistance of a bilingual dictionary extracted from the Catalan--English Apertium rule-based MT system.\footnote{\url{http://sourceforge.net/projects/apertium/files/apertium-en-ca/0.9.3/}}
Following empirical observations, we keep aligned sentences with confidence scores higher than 0.3 and 0.5 for the training and test sets, respectively.

Subsequently, all the datasets are truecased and normalised in terms of punctuation with Moses' scripts.
Finally, in the  parallel training data we discard sentence pairs where either of the sides has fewer than 1 or more than 80 tokens.

\subsubsection{MT Toolkits and Evaluation}\label{s:toolkits_and_eval}

PBSMT systems are trained with version 3 of the Moses toolkit~\citep{Koehn:2007:MOS:1557769.1557821},
while
NMT systems are trained with Nematus~\citep{nematus}.\footnote{\url{https://github.com/rsennrich/nematus}}
For both paradigms default settings are used, unless mentioned otherwise in the description of the experiments (see Sections \ref{s:pbmt} and \ref{s:nmt} for PBSMT and NMT, respectively).


Automatic evaluation is carried out with the BLEU metric and is case-insensitive.
Multi-bleu as implemented in Moses 3.0 is used for evaluating the development set while mteval (13a) is used to evaluate the test set.
Statistical significance of the difference between systems is computed with paired bootstrap resampling~\citep{koehn2004statistical} ($p\le0.01$, $1\,000$ iterations).\footnote{\url{http://www.cs.cmu.edu/~ark/MT/paired_bootstrap_v13a.tar.gz}}
Human evaluation is ranked-based and is performed with the Appraise tool~\citep{mtm12_appraise}.\footnote{\url{https://github.com/cfedermann/Appraise}}


\section{MT Systems}\label{s:mt_systems}

\subsection{PBSMT System}\label{s:pbmt}


The PBSMT system is trained on both the in-domain and out-of-domain parallel datasets by means of linear interpolation~\citep{Sennrich:2012:PMT:2380816.2380881} and uses three reordering models (lexical- and phrase-based and hierarchical).
In addition, the system makes use of additional feature functions based on the operation sequence model (OSM)~\citep{durrani2011joint} and language models based not only on surface $n$-grams but also on continuous space $n$-grams (NPLM)~\citep{vaswani2013decoding}.
The OSM and NPLM models are built on the in-domain parallel data (both sides in the case of OSM and only the target side for NPLM).
The vocabulary size for NPLM is set to 100,000.
Surface-form $n$-gram language models are built on the in-domain and out-domain datasets with KenLM~\citep{heafield2011kenlm} and then linearly interpolated with SRILM~\citep{stolcke2002srilm}.
Tuning is carried out with batch MIRA~\citep{cherry2012batch}.

During development we tuned PBSMT systems using different subsets of the components previously introduced in order to assess their effect on translation quality as measured by the BLEU evaluation metric.
Table \ref{t:dev_smt} shows the results, where we start with a baseline trained on in-domain data (in) both for the translation model (TM) and the language model (LM) and we measure the effect of the following:
\begin{itemize}
\item Adding NPLM, both using 4-grams and 5-grams, which results in absolute improvements of 0.57 and 0.75 BLEU points, respectively.
\item Adding OSM ($+0.4$).
\item Adding linearly interpolated out-domain data both for the TM and the LM ($+0.14$).
\end{itemize}

\begin{table}[htbp]
\begin{center}
\begin{tabular}{llllr}
\bf TM & \bf LM & \bf OSM & \bf NPLM & \bf BLEU\\
\hline
in		&in			&-		&-			&0.3344\\
in		&in			&- 		&4-gram	&0.3401\\
in		&in			&- 		&5-gram	&0.3419\\
in		&in			&y		&5-gram	&0.3459\\
inIout	&inIout		&y		&5-gram 	&0.3473\\
\hline
\end{tabular}
\end{center}
\caption{Performance of different configurations of the PBSMT system on the development set.}
\label{t:dev_smt}
\end{table}

\subsection{NMT System}\label{s:nmt}

Due to the lack of established domain adaptation techniques for NMT at the time when this system was built, our NMT system was trained solely on in-domain data.
Specifically, we trained our NMT system on the concatenation of the parallel in-domain training data and a synthetic corpus obtained by machine translating the Catalan in-domain monolingual training data into English.

We use additional parallel data in which the source side is synthetic (machine-translated from the target language), 
as this has been reported to be a successful way of integrating target-language monolingual data into NMT~\citep{sennrich2015b} (see also footnote 25 in Way's chapter in this volume).
The in-domain monolingual training data for Catalan is translated into English by means of a Catalan-to-English PBSMT system built for this purpose. This PBSMT system is based on the PBSMT system described in Section \ref{s:pbmt}. 
Aside from reversing the translation direction, this PBSMT system is trained on the same datasets and has the same components, except for the following, which are not used: out-of-domain training data (both parallel and monolingual) and NPLM.
The reason not to use these components has to do with an efficiency versus translation quality trade-off; this system needs to be fast as it is used to translate over 5 million sentences (i.e. the in-domain monolingual training data for Catalan), and taking the example of NPLM, this is a rather computationally expensive component to run. 

We limit the source and target vocabularies to the 50,000 most frequent tokens in the respective sides of the training data.
Training is then run until convergence, with models being saved every 3 hours.\footnote{Training is performed on an NVIDIA Tesla K20X GPU.}
Each model is evaluated on the development set using BLEU in order to track performance over training time and find out when the training reaches convergence.

\begin{figure}[htbp]
\includegraphics[width=\textwidth]{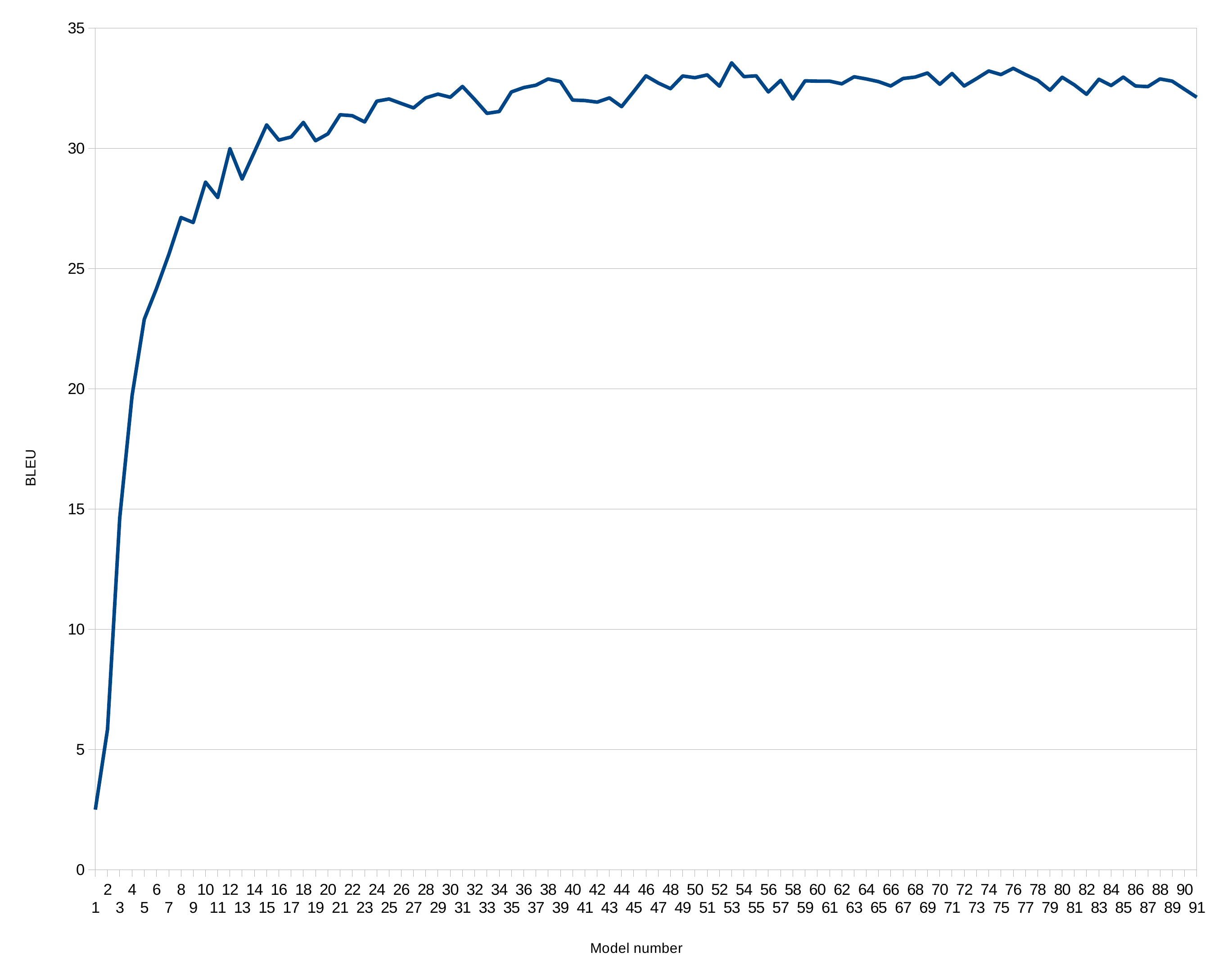}
\caption{BLEU scores obtained by the NMT models on the development set. Each model is saved every 3 hours.}
\label{f:nmt_training}
\end{figure}

Figure \ref{f:nmt_training} shows the results.
We can observe that performance increases very quickly in the first iterations, going from 0.0251 BLEU points for model 1 (i.e. after 3 hours of training) to 0.2999 for model 12 (i.e. after 36 hours), after which it grows slowly to reach its maximum (0.3356) for model 53 and then plateaus.

We select the 4 models with the highest BLEU scores.
These are, in descending order,
53 (0.3356 points), 76 (0.3333), 74 (0.3322) and 69 (0.3314).
We trained these models for 12 hours with the embeddings frozen (i.e. the whole network keeps being trained except the first layer (embeddings) which is fixed).
We then evaluate ensembles of these 4 models `as is' as well as with the additional training for 12
hours with fixed embeddings.
Their BLEU scores are 0.3561 (2.05 points absolute higher than the best individual system, a 6.1\% relative improvement) and 0.3555 (1.99 points absolute higher than the best individual system, 5.9\% relative), respectively.
In other words, ensembling led to a substantial improvement, but fixing embeddings -- reported to provide further improvements in several experiments in the literature -- did not increase  performance in our setup.

Subsequently, we tried to improve upon this NMT system by implementing the following two functionalities:
\begin{enumerate}
\item Using subwords rather than words as translation units. Specifically, we segmented the training data into characters and performed 90,000 operations jointly on both the source and target languages~\citep{sennrich-haddow-birch:2016:P16-12}. These operations iteratively join the most frequent pair of segments. This results in a score of 0.3689 (1.28 points absolute higher than the initial NMT ensemble, a 3.6\% relative improvement).

\item Producing an $n$-best list and reranking it with a right-to-left NMT system~\citep{sennrich-haddow-birch:2016:WMT}. We trained a so-called right-to-left system, with the same settings and the only difference being that the target sentences of the training data are reversed at the word level. We then produced an $n$-best list containing the top-50 translations with the previous model and re-ranked it with the right-to-left model. This leads to a BLEU score of 0.3948 (2.59 points higher than the previous system, a 7\% relative improvement), and almost 6 BLEU points better (a 17.6\% relative improvement) than the best individual system.
\end{enumerate}

Due to the fact that we use the same dataset for development in the PBSMT and NMT paradigms, we are able to compare their results.
When doing so, however, one should take into account that any such comparison would be unfair in favour of PBSMT.
This is because in the development of PBSMT the system is optimising its log-linear weights to obtain the highest performance on the development set.
Conversely, in the development of NMT we use the development set for validation, i.e. the system is not optimised on the development set.
Despite the bias towards PBSMT, we observe that the score obtained by the best NMT system (0.3948, ensemble, using subword units and re-ranked with a right-to-left model) is notably higher (4.75 points, a 13.7\% relative improvement) than the score achieved by the best PBSMT system (0.3473, all components, see Table \ref{t:dev_smt}).

\section{Evaluation}\label{s:evaluation}

\subsection{Automatic Evaluation}
As previously mentioned in Section~\ref{s:toolkits_and_eval}, we automatically evaluate the MT systems  using the BLEU metric.
Table \ref{t:results} shows the BLEU scores obtained for each novel in the test set with both PBSMT and NMT.
The results across the different novels show a very high degree of variability, indicating that fiction is far from being a monolithic domain.
In fact scores go from a low of 0.1611 (PBSMT for {\it Ulysses}) to the highest of 0.3892 (NMT for {\it Harry Potter \#7}), which more than doubles the first figure.

As for the performance obtained by the two paradigms that we compare in this chapter, NMT beats PBSMT by a statistically significant margin for all novels. 
On average, NMT outperforms PBSMT by 10.67\% relative and 3 points absolute.
The improvement brought about by NMT compared to PBSMT varies widely depending on the book, going from 3.11\% (Auster's {\it Sunset Park}) to 14\% (Collins' {\it Hunger Games \#3}).


\begin{table}[h]
\begin{center}
\begin{tabular}{lrrr}
\bf Novel & \bf PBSMT & \bf NMT & \bf Relative improvement\\
\hline
Auster's {\it Sunset Park} (2010)					&0.3735 &0.3851 &3.11\%\\
Collins' {\it Hunger Games \#3} (2010) 				&0.3322 &0.3787 &  14.00\%\\
Golding's {\it Lord of the Flies} (1954)			&0.2196 &0.2451 &  11.61\%\\
Hemingway's {\it The Old Man and the Sea} (1952)	&0.2559 &0.2829 &  10.55\%\\
Highsmith's {\it Ripley Under Water} (1991) 		&0.2485 &0.2762 &  11.15\%\\
Hosseini's {\it A Thousand Splendid Suns} (2007) 		&0.3422 &0.3715 &  8.56\%\\
Joyce's {\it Ulysses} (1922) 						&0.1611 &0.1794 &  11.36\%\\ 
Kerouac's {\it On the Road} (1957) 					&0.3248 &0.3572 &  9.98\%\\
Orwell's {\it 1984} (1949) 							&0.2978 &0.3306 &  11.01\%\\
Rowling's {\it Harry Potter \#7} (2007) 			&0.3558 &0.3892 &  9.39\%\\
Salinger's {\it The Catcher in the Rye} (1951) 		&0.3255 &0.3695 &  13.52\%\\
Tolkien's {\it The Lord of the Rings \#3} (1955) 		&0.2537 &0.2888 &  13.84\%\\
\hline
Average												&0.2909 &0.3212 &  10.67\%\\
\end{tabular}
\end{center}
\caption{BLEU scores obtained by PBSMT and NMT for each of the books that make up the test set.
NMT outperforms PBSMT by a statistically significant margin ($p<0.01$) on all the books.}
\label{t:results}
\end{table}

\subsubsection{Analysis}

We performed a set of additional analyses in order to obtain further insights from the output translations and, especially to try to find the reason why NMT, while outperforming PBSMT for all the novels, does so by rather diverging margins (from a minimum of 3.11\% to a maximum of 14\%, see Table~\ref{t:results}).

More specifically, we considered three characteristics of the source-side of each novel in the test set (lexical richness, novelty with respect to the training data, and average sentence length) and studied whether any of these features correlates to some extent with the performance of the PBSMT and NMT systems and/or with the relative improvement of NMT over PBSMT.
The motivation to use these three features is as follows:
\begin{itemize}
\item Lexical richness has been already studied in relation to NMT, and there are indications that this MT paradigm has ``an edge especially on lexically rich texts"~\citep{bentivogli-EtAl:2016:EMNLP2016}.
\item There is evidence that NMT's performance degrades with sentence length~\citep{E17-1100}.
\item Despite, to the best of our knowledge, the lack of empirical evidence, it is still the perceived wisdom that NMT is better at generalising than PBSMT, and so it should perform better than the latter especially on data that is unrelated to the training data.
\end{itemize}

\paragraph{\bf Lexical richness}

We use type-token ratio (TTR) as a proxy to measure lexical richness.
The higher the ratio, the less repetitive the text and hence it can be considered lexically more varied, and thus richer.
To measure this we calculate the TTR on the source side of each novel.
As they have different sizes, we calculate the TTR for each novel on a random subset of sentences that amount to approximately $n$ words, $n$ being $20,000$, a slightly lower number to the number of words contained in the smallest novel in our dataset, {\it The Old Man and the Sea} with $24,233$ words.

\paragraph{\bf Sentence length}

We measure the average sentence length of each novel as the ratio between its total number of tokens and its number of sentences.
Both these values were reported in Table \ref{t:testset}.

\paragraph{\bf Novelty with respect to the training data}

We use $n$-gram overlap to measure the novelty of a novel with respect to the training data.
Concretely, we consider the unique $n$-grams ($n=4$) in the parallel training data and in each novel, and calculate the overlap as the ratio between the size of the intersection and the number of unique $n$-grams in the training set.
The higher the overlap, the less novelty that the novel presents with respect to the training data.
As in the analysis concerning lexical richness, we consider 20,000 words from randomly selected sentences for each novel.

\paragraph{\bf Results}

Table \ref{t:analysis_raw} shows the values for each novel and for each of the three features analysed.
We can clearly observe an outlier in the data for all the three variables reported.
{\it Ulysses} has the highest TTR by far at 0.276 and is also the novel with the lowest overlap by a wide margin (0.216).
As for sentence length, the value for {\it Sunset Park} (32.434) is over 10 points higher than the value for any other novel.

\begin{table}[htbp]
\begin{center}
\begin{tabular}{lccc}
\hline
\multirow{2}{*}{\bf Novel} &\multirow{2}{*}{\bf TTR} & \bf avg. sentence &\multirow{2}{*}{\bf Overlap}\\
&	& \bf length&\\
\hline
Auster's {\it Sunset Park} (2010)		&0.1865	&\bf 32.434		&0.368\\
Collins' {\it Hunger Games \#3} (2010)		&0.1716	&14.177		&0.393\\
Golding's {\it Lord of the Flies} (1954)	&0.1368	&12.442		&0.370\\
Hemingway's {\it The Old Man and the Sea} (1952)&0.1041	&16.587		&0.371\\
Highsmith's {\it Ripley Under Water} (1991)	&0.1492	&14.101		&0.404\\
Hosseini's {\it A Thousand Splendid Suns} (2007)	&0.1840	&14.765		&0.377\\
Joyce's {\it Ulysses} (1922)			&\bf 0.2761	&12.185		&\bf 0.216\\
Kerouac's {\it On the Road} (1957)		&0.1765	&17.902		&0.335\\
Orwell's {\it 1984} (1949)			&0.1831	&17.325		&0.343\\
Rowling's {\it Harry Potter \#7} (2007)		&0.1665	&17.031		&0.433\\
Salinger's {\it The Catcher in the Rye} (1951)	&0.1040	&13.900		&0.448\\
Tolkien's {\it The Lord of the Rings \#3} (1955)	&0.1436	&18.497		&0.368\\
\hline
\end{tabular}
\end{center}
\caption{TTR, average sentence length and 4-gram overlap for the source side of the 12 novels that make up the test set.
The highest TTR and average sentence length as well as the lowest $n$-gram overlap values are shown in bold.
}
\label{t:analysis_raw}
\end{table}

\begin{figure}[htbp]
\includegraphics[width=0.8\textwidth]{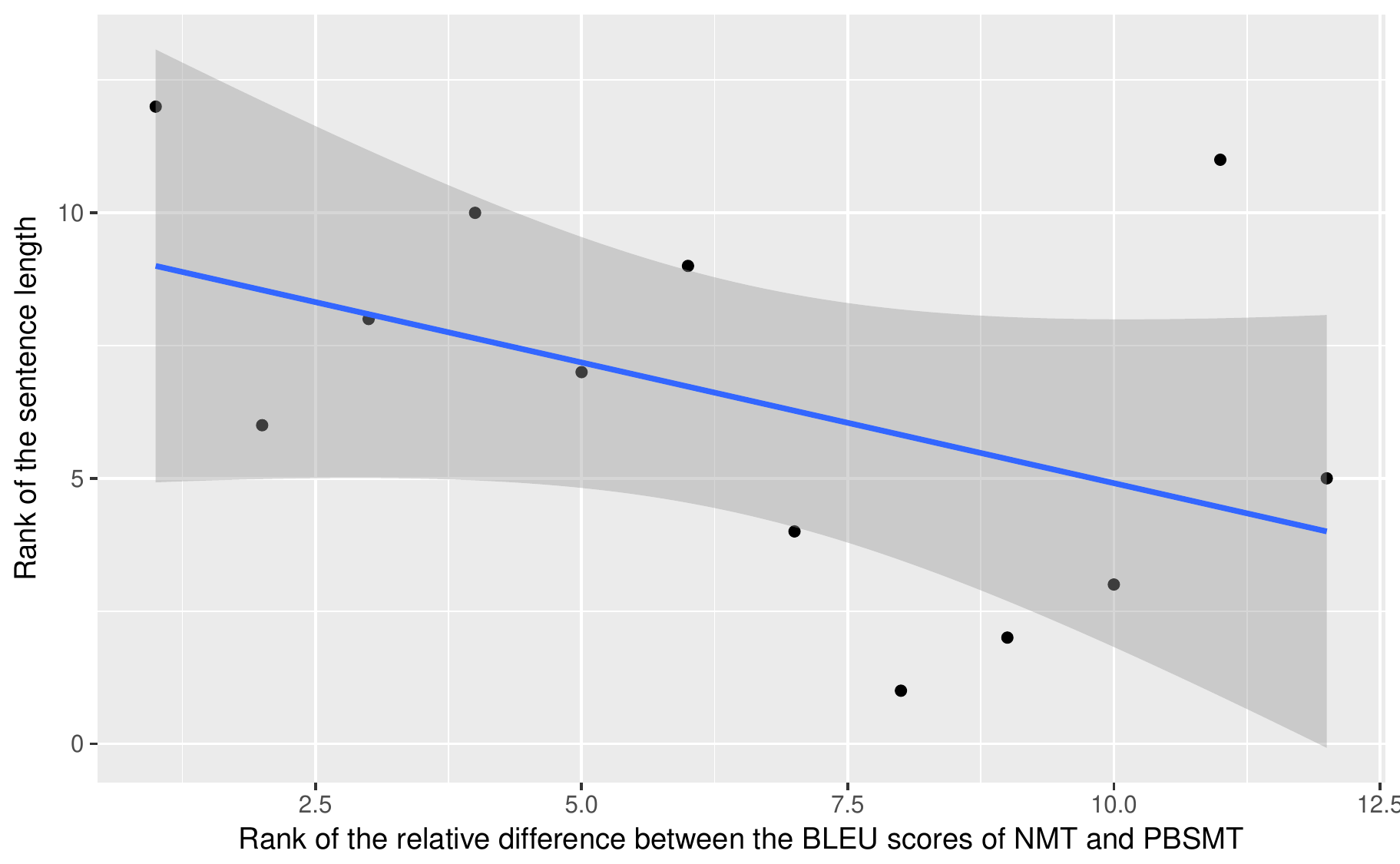}
\caption{Spearman correlation between the relative difference between the BLEU scores of the NMT and PBSMT systems and sentence length.}
\label{f:bleureldiff_sentlength}
\end{figure}

\begin{figure}[htbp]
\includegraphics[width=0.8\textwidth]{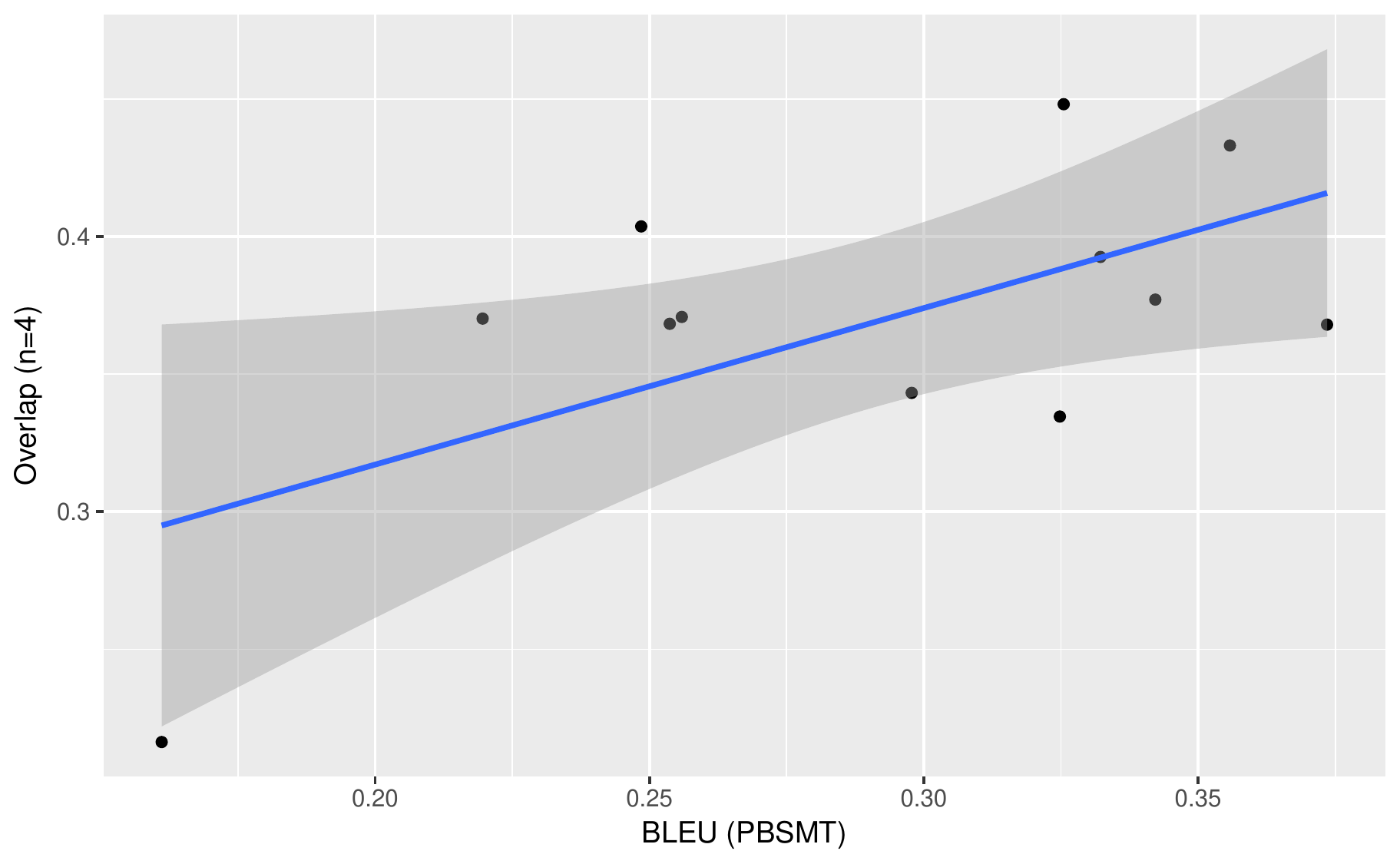}
\caption{Pearson correlation between the the BLEU score of the PBSMT system and 4-gram overlap.}
\label{f:bleupbmt_overlap4}
\end{figure}

\begin{figure}[htbp]
\includegraphics[width=0.8\textwidth]{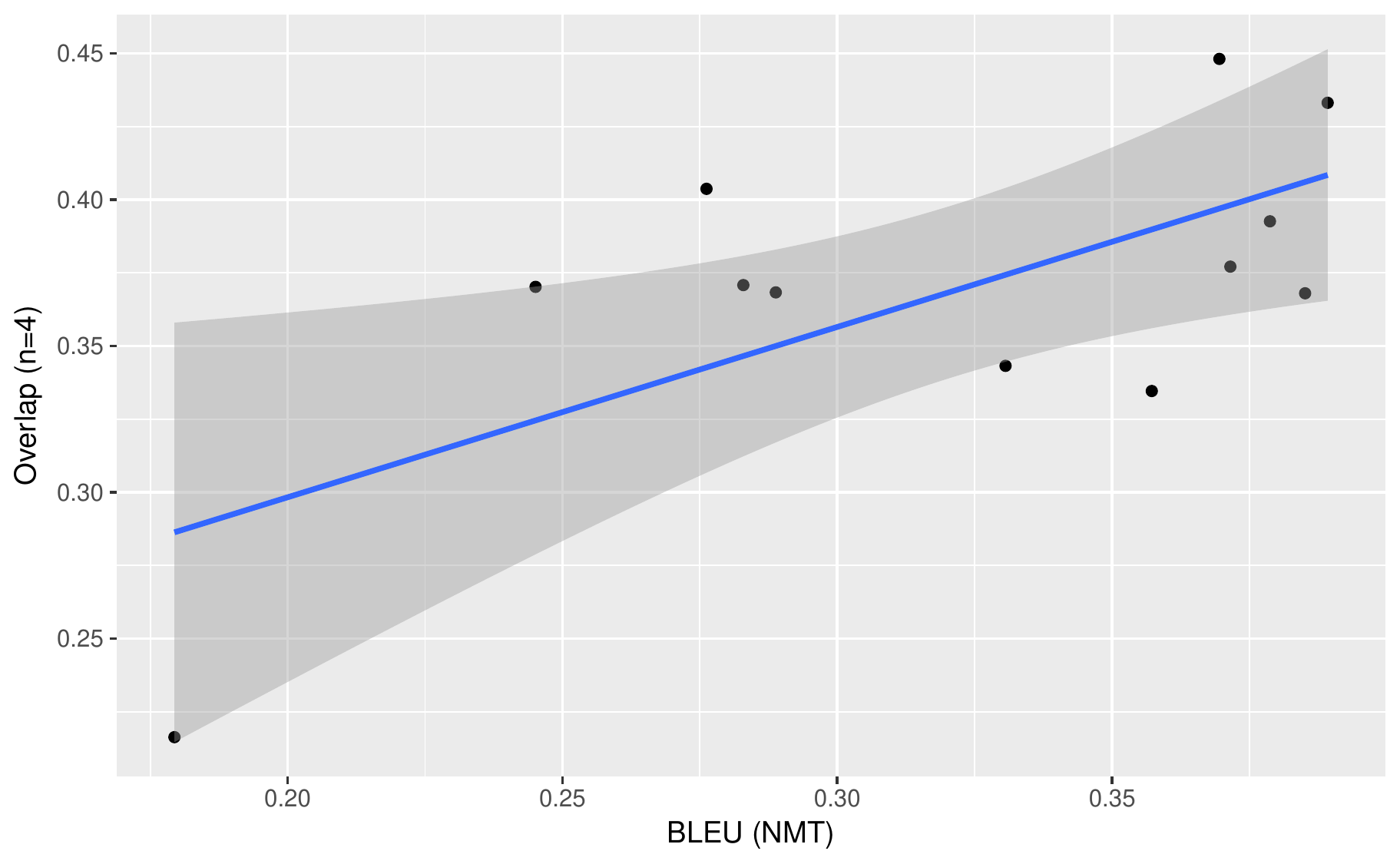}
\caption{Pearson correlation between the the BLEU score of the NMT system and 4-gram overlap.}
\label{f:bleunmt_overlap4}
\end{figure}

Table \ref{t:analysis_correlations} shows the significant correlations between the BLEU scores (for PBSMT, NMT and the relative difference between both, see Table \ref{t:results}) and the three variables analysed (TTR, average sentence length and $n$-gram overlap).
Each of the significant correlations is then plotted, including its regression line and its 95\% confidence region, in Figures \ref{f:bleureldiff_sentlength}, \ref{f:bleupbmt_overlap4} and \ref{f:bleunmt_overlap4}.

\begin{table}[htbp]
\begin{center}
\begin{tabular}{lccc}
\hline
\multirow{2}{*}{\bf BLEU} &\multirow{2}{*}{\bf TTR} & \bf avg. sentence &\multirow{2}{*}{\bf Overlap}\\
&	& \bf length		&\\
\hline
PBSMT		&- 			& - 							& $r$=0.62, $p<0.05$\\
NMT		&-			& - 							& $r$=0.66, $p<0.01$\\
rel. diff		&- 			& $\rho$=-0.45\tablefootnote{A significant parametric Pearson correlation was found ($r=-0.78$, $p<0.01$) but the assumption that both variables come from a bivariate normal distribution was not met, hence the reason why a non-parametric Spearman correlation is shown instead.}, $p$=0.07	& -\\
\hline
\end{tabular}
\end{center}
\caption{Correlations between the BLEU scores for NMT, PBSMT and their relative difference and the other metrics considered (TTR, average sentence length and 4-gram overlap) for the 12 novels that make up the test set.
Empty cells mean that no significant correlation was found.
}
\label{t:analysis_correlations}
\end{table}

While \cite{bentivogli-EtAl:2016:EMNLP2016} found a moderate correlation ($r=0.73$) between TTR and the gains by NMT over PBSMT (measured with mTER -- multi-reference TER~\citep{snover} -- on transcribed speeches), there is no significant correlation in our setup.

With respect to sentence length (see Figure~\ref{f:bleureldiff_sentlength}), we observe a negative correlation ($\rho=-0.45$), meaning that the relative improvement of NMT over PBSMT decreases with sentence length.
This corroborates the findings in previous work~\citep{E17-1100} and appears to be the main reason behind the low relative improvement that NMT achieved for {\it Sunset Park} (see Table \ref{t:results}), as the average sentence length for this novel is very long compared to all the other novels in our test set (see Table \ref{t:analysis_raw}).

Finally, we found significant positive correlations between the performance of both PBSMT and NMT and $n$-gram overlap, the Pearson correlation coefficient being $r=0.62$ (see Figure~\ref{f:bleupbmt_overlap4}) and $r=0.66$ (see Figure~\ref{f:bleunmt_overlap4}), respectively.
This matches the intuition that the performance of a statistical MT system should be better the more the test set resembles the training data.
That said, we did not find significant correlations between the relative improvement of NMT over PBSMT and overlap.
Thus, the perceived wisdom that the more unrelated the data to be translated from the test set the wider the gap between NMT and PBSMT, does not hold for our setup.

\subsection{Human Evaluation}

\textcolor{blue}{We also conducted a manual evaluation, in order to gain further insights.}
A common procedure (e.g. conducted in the annual MT shared task at WMT)\footnote{e.g. \url{http://www.statmt.org/wmt17/translation-task.html}} consists of ranking the translations produced by different MT systems (see also Section 3.4 in Catilho et al. in this volume).
Given the source and target sides of the reference (human) translations, and two or more outputs from MT systems, these outputs are ranked according to their quality, e.g. in terms of adequacy and/or fluency.

In our experiment, we are not only interested in comparing two MT systems -- PBSMT and NMT -- to each other, but also with respect to the human reference translation.
Hence, we conduct the rank-based manual evaluation in a slightly modified setting; we do not provide the target of the reference translation as reference but as one of the translations to be ranked.
The evaluator thus is provided with the source-side of the reference and three translations, one being the human translation and the other two the translations produced by the PBSMT and NMT systems.
The evaluator of course does not know which is which. Moreover, in order to avoid any bias with respect to MT, they are not told whether the translations are human or automatic.

This human evaluation is conducted for three of the books used in the automatic evaluation: Orwell's {\it 1984}, Rowling's {\it Harry Potter \#7} and Salinger's {\it The Catcher in the Rye}.
For each of these books, the sentences in 10 randomly selected passages were ranked.
Each passage is made of 10 contiguous sentences, the motivation being to provide the annotator with context beyond the sentence level. Therefore, sentences 1 to 10 (passage 1) are contiguous in the book, then there is a jump to a second passage contained in sentences 11 to 20, and so forth.

All the annotations were carried out by two native Catalan speakers with an advanced level of English. They both have a background in linguistics but no in-depth knowledge of statistical MT (again, to avoid any bias with respect to MT).
Comprehensive instructions were provided to the evaluators in their native language, in order to minimise ambiguity and thus foster high inter-annotator agreement. 
Here we reproduce the translation into English of the evaluation instructions: 

Given three translations, the task is to rank them:
\begin{itemize}
\item Rank a translation $A$ higher (rank 1) than a translation $B$ (rank 2), if the first translation is better than the second.
\item  Rank two translations $A$ and $B$ equally (rank 1 for both $A$ and $B$), if both have an equivalent quality.
\item  Use the highest rank possible, e.g. if there are three translations $A$, $B$ and $C$,
and the quality of $A$ and $B$ is equivalent and both are better than $C$, then
they should be ranked as follows: $A$ = rank 1, $B$ = rank 1, $C$ = rank 2. Do NOT use lower rankings, e.g.: $A$ = rank 2, $B$ = rank 2, $C$ = rank 3.
\end{itemize}

Please follow the following guidelines to decide that a translation is better than another:
\begin{itemize}
\item  Above anything else: the meaning of the original is understood, all the information is preserved and, if possible, the translation sounds natural.
\item  If all translations preserve the meaning to a similar extent, you might compare the number of errors (e.g. lexical, syntax, etc) in each translation.
\item If for a given set of translations, you cannot decide how to rank them, you can skip that set by pressing the button ``flag error''.
\end{itemize}

\begin{figure}[htbp]
\includegraphics[width=\textwidth]{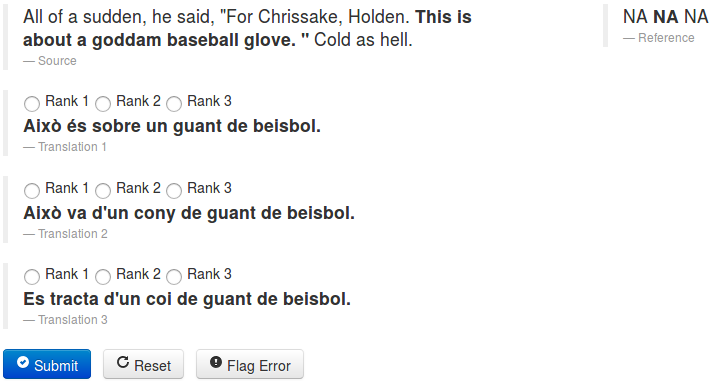}
\caption{Snapshot from the manual evaluation process.}
\label{f:appraise_snapshot}
\end{figure}

Figure \ref{f:appraise_snapshot} shows a snapshot of the manual evaluation process. In this example the annotator is asked to rank three translations for the second sentence from the first passage of Salinger's {\it The Catcher in the Rye}.

\subsubsection{Inter-annotator Agreement}
The inter-annotator agreement in terms of Fleiss' Kappa~\citep{fleiss1971mns} is 0.22 for Orwell, 0.18 for Rowling and 0.38 for Salinger.
The values for Orwell and Salinger fall in the band of fair agreement $[0.21, 0.4]$~\citep{LandisEA77} while that for Rowling is at the higher end of slight agreement $[0.01, 0.2]$.
For the sake of comparison, the average inter-annotator agreement at WMT for the closest language direction to ours (English-to-French) over the last four editions in which that language direction was considered is 0.29, see Table 4 in \cite{bojar-EtAl:2016:WMT1}.

\subsubsection{Pairwise Rankings}\label{s:pairwise_rankings}
From the sets of annotations (rankings between three translations), we extract all the pairwise rankings, i.e. the rankings for each pair of translations.
Given two translations $A$ and $B$, the pairwise ranking will be $A>B$ if translation $A$ was ranked higher  than $B$, $A<B$ if $A$ was ranked lower than $B$, and $A=B$ if both were ranked equally.

It is worth mentioning that while the PBSMT translations consistently cover the source sentences, this is not always the case for the other two translations.
NMT has a tendency towards omission errors~\citep{pbml-2017-mqm-nmt-pbmt}. 
The human translation sometimes does not cover the source sentence fully either.
This may be due to a choice of the translator, e.g. to translate the sentence in a way that diverges notably from the source.
There are also some cases where the human translation is misaligned\footnote{As mentioned in Section~\ref{s:preprocessing}, the source novels and their human translations were sentence-aligned automatically. 
The empirically set confidence threshold results in most alignments being correct, but some are erroneous.
} and so it is unrelated to the source sentence.
Most cases in which the human translation is ranked lower than MT (PBSMT or NMT) are due to either of these two reasons.
It is clearly unjustifiable to rank the human translation lower than MT in these cases,
so we remove these pairwise rankings, i.e. $A<B$ where $A$ is the human translation and $B$ corresponds to the translation produced by either MT system.\footnote{While the majority of HT$<$MT cases are unjustified, not all of them are. By removing these rankings, the results are slightly biased in favour of HT and thus overly conservative with respect to the potential of MT.}

\begin{figure}[htbp]
\includegraphics[width=0.82\textwidth]{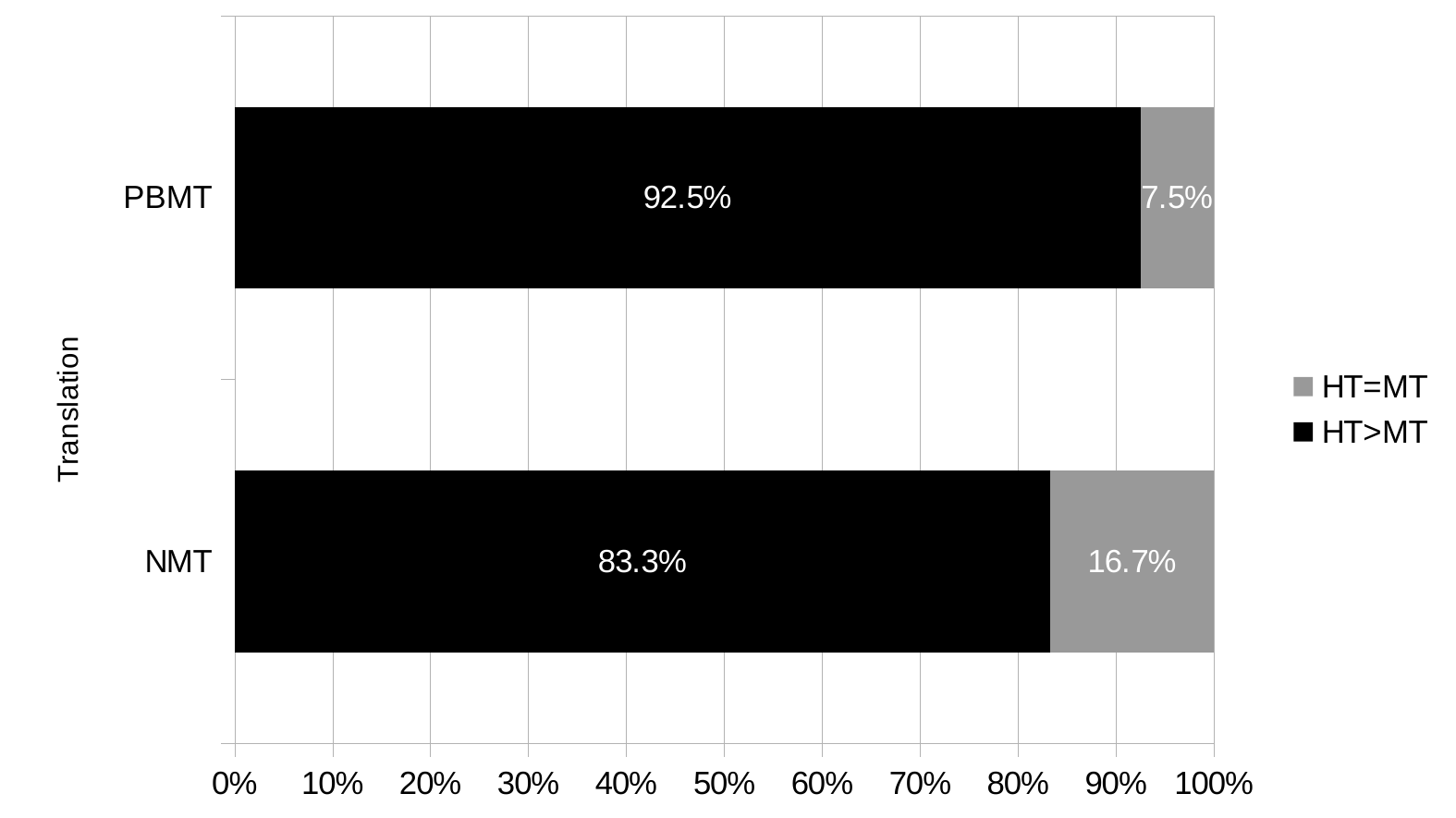}
\caption{Pairwise rankings between HT and MT for Orwell's {\it 1984}.}
\label{f:ht_mt_orwell}
\end{figure}

\begin{figure}[htbp]
\includegraphics[width=0.82\textwidth]{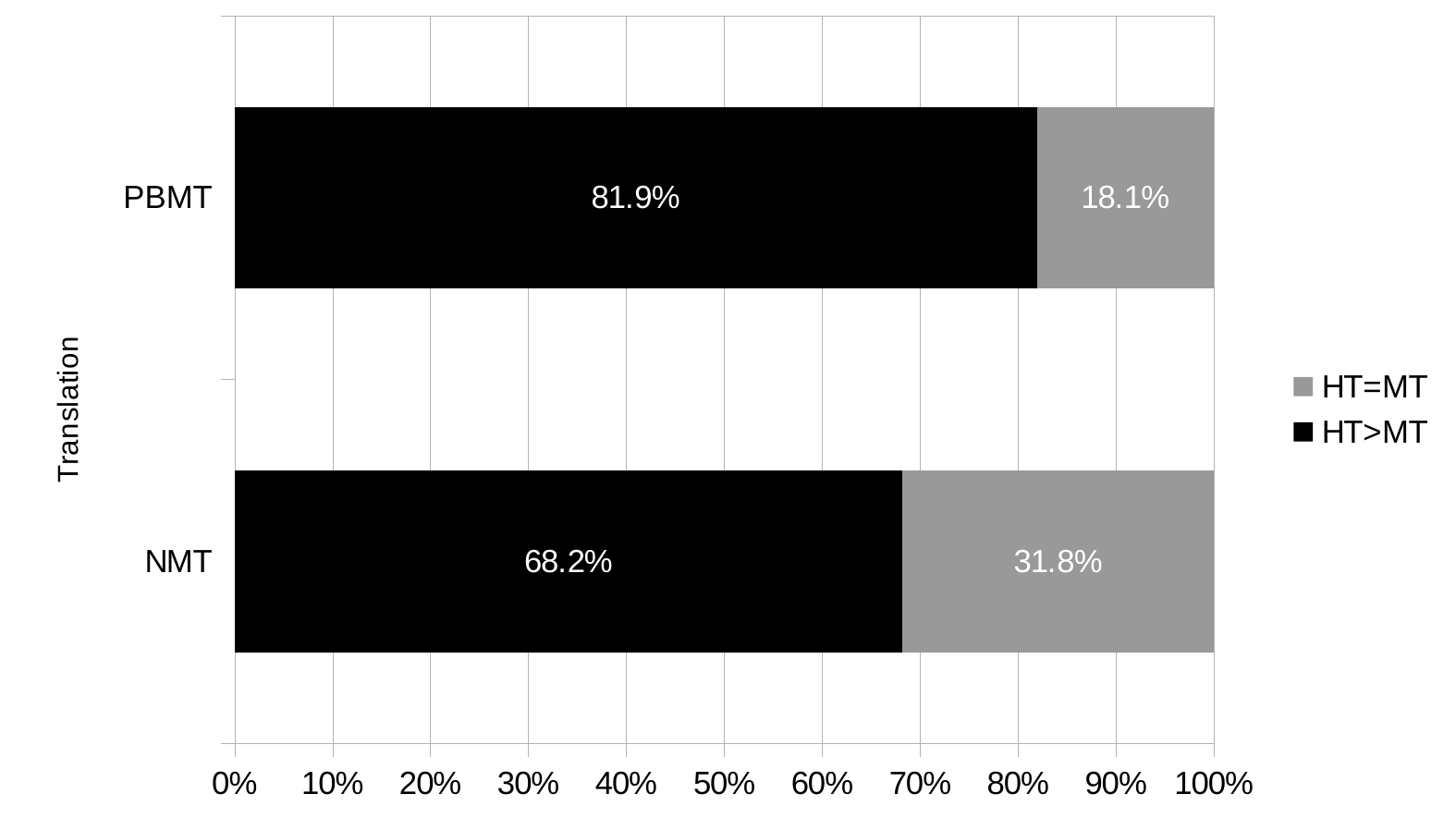}
\caption{Pairwise rankings between HT and MT for Rowling's {\it Harry Potter \#7}.}
\label{f:ht_mt_rowling}
\end{figure}

\begin{figure}[htbp]
\includegraphics[width=0.82\textwidth]{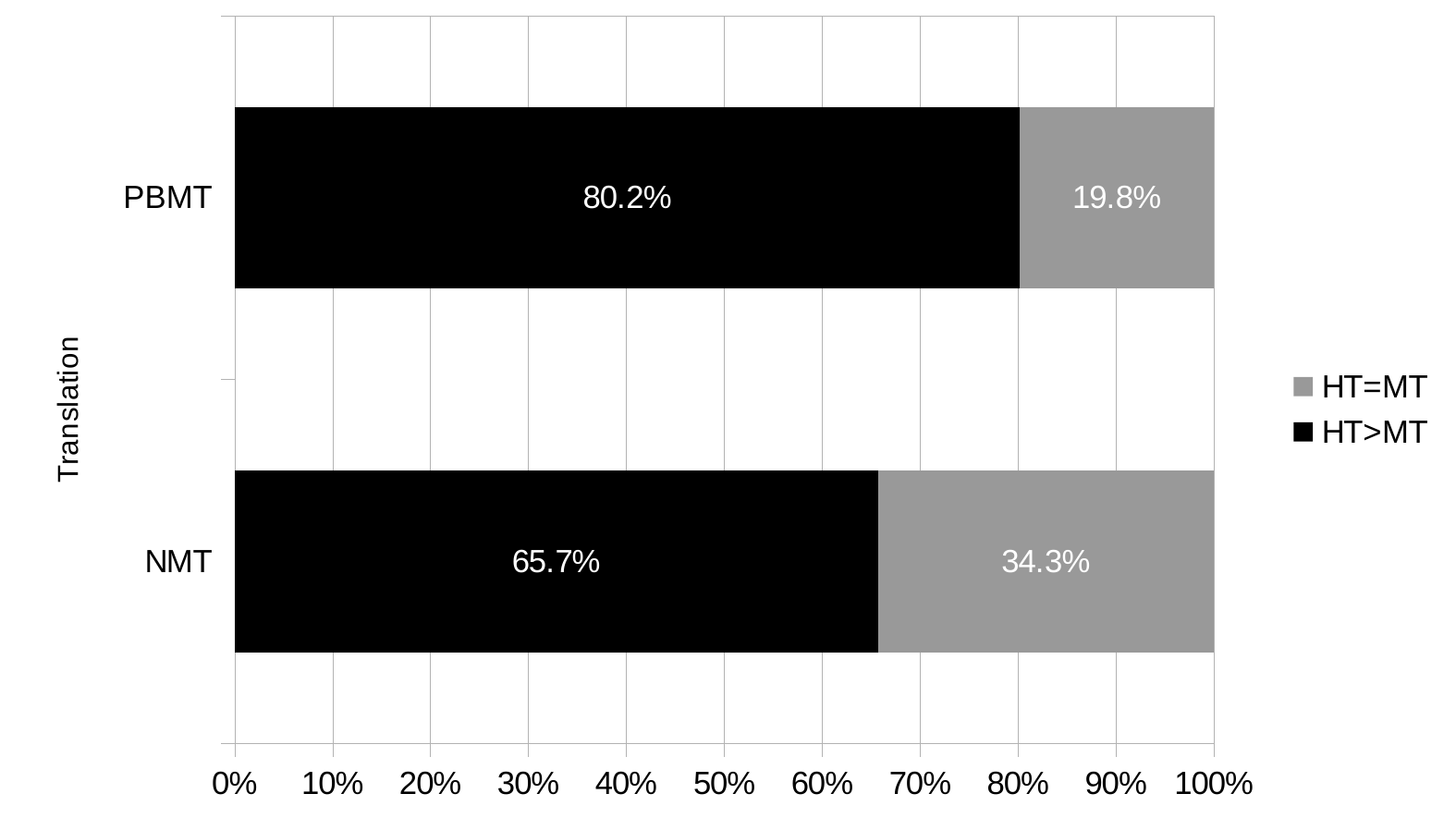}
\caption{Pairwise rankings between HT and MT for Salinger's {\it The Catcher in the Rye}.}
\label{f:ht_mt_salinger}
\end{figure}

Figures \ref{f:ht_mt_orwell}, \ref{f:ht_mt_rowling} and \ref{f:ht_mt_salinger}  show the pairwise rankings between each MT system and the human translation (HT) for Orwell's, Rowling's and Salinger's books.
In all three books, the percentage of sentences where the annotators perceive the MT translation to be of equivalent quality to the human translation is considerably higher for NMT compared to PBSMT: 16.7\% vs 7.5\% for Orwell's, 31.8\% vs 18.1\% for Rowling's and 34.3\% vs 19.8\% for Salinger's.
In other words, if NMT translations were to be used to assist a professional translator (e.g. by means of post-editing), then around one third of the sentences for Rowling's and Salinger's and one sixth for Orwell's would not need any correction.

\begin{figure}[htbp]
\includegraphics[width=0.82\textwidth]{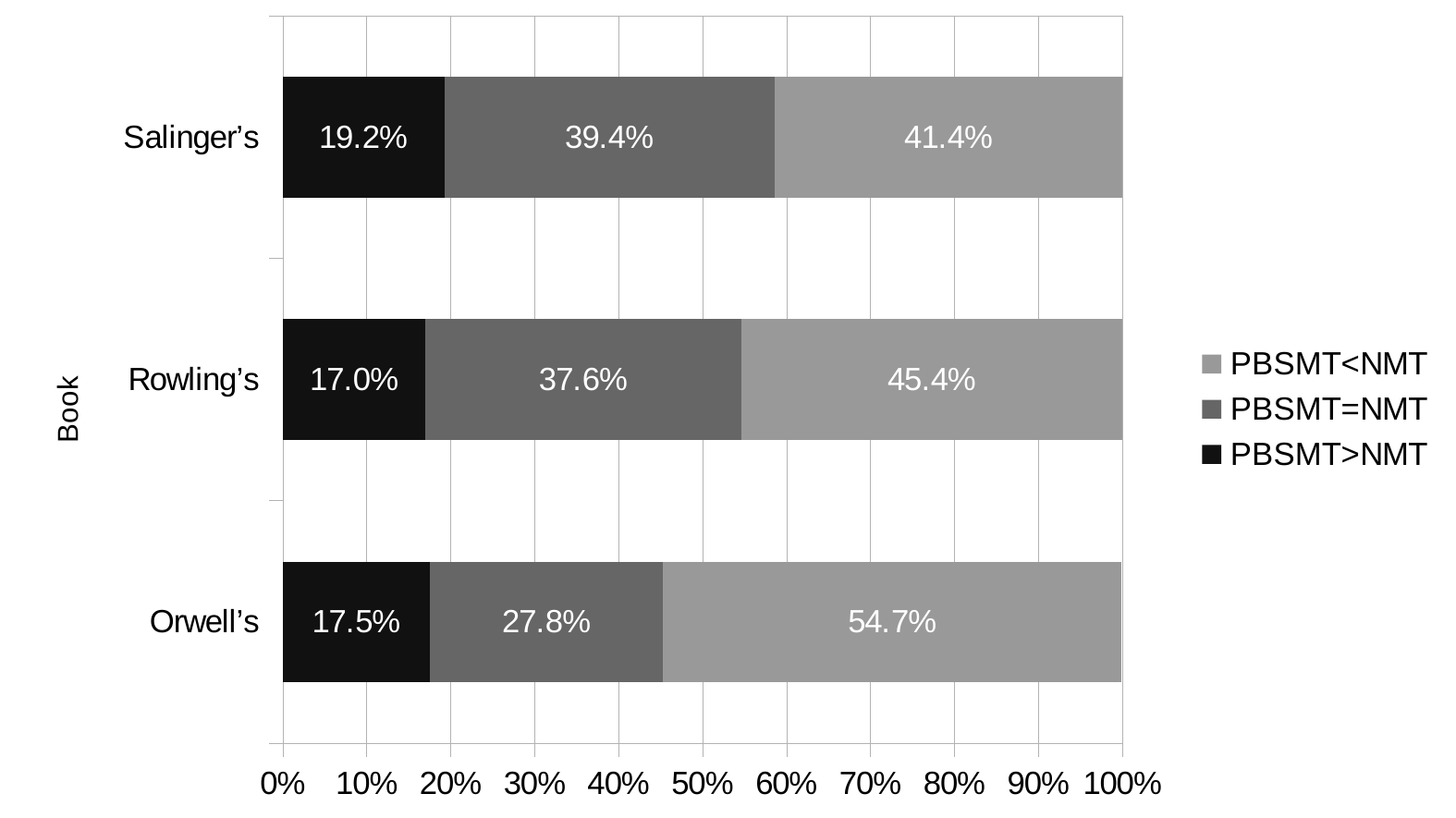}
\caption{Pairwise rankings between PBSMT and NMT.}
\label{f:pbmt_nmt}
\end{figure}

Having looked at pairwise rankings between MT and human translations,
we move our attention now to the pairwise rankings between the two types of MT systems.
The results for all three books are depicted in Figure~\ref{f:pbmt_nmt}.
In all the books the trends are similar.
The biggest chunk (41.4\%, 54.7\%) corresponds to cases where NMT translations are ranked higher than PBSMT's (PBSMT$<$NMT).
The second relates to translations by both systems which are ranked equally (27.8\%, 39.4\%), PBSMT$=$NMT).
Finally, the smallest chunk (less than 20\% in all three books) signifies translations for which PBSMT is ranked higher than NMT (PBSMT$>$NMT).

%
%
%
%
%
%
%


\subsubsection{Overall Human Scores}


In addition to the pairwise rankings,
we derive an overall score for each translation type (HT, NMT and PBSMT) and novel based on the rankings.
To this end we use the TrueSkill method adapted to MT evaluation~\citep{sakaguchi-post-vandurme:2014:W14-33} 
following its usage at WMT15.\footnote{\url{https://github.com/mjpost/wmt15}}
Namely, we run 1,000 iterations of the rankings recorded with Appraise followed by clustering ($p<0.05$).

\begin{figure}[htbp]
\includegraphics[width=1.0\textwidth]{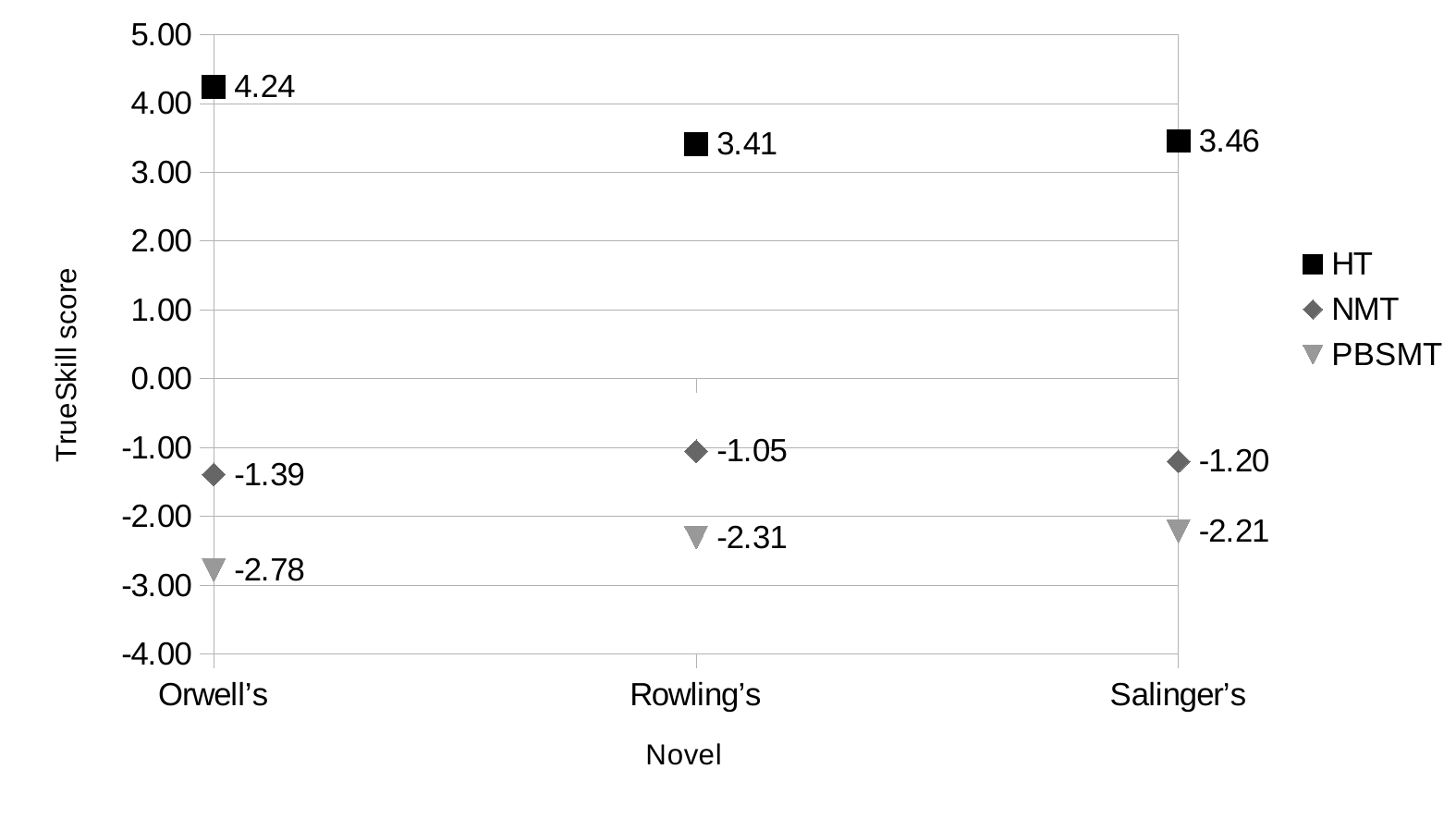}
\caption{Overall human evaluation scores with TrueSkill.}
\label{f:trueskill}
\end{figure}

Figure~\ref{f:trueskill} depicts the results.
For all the three books considered, all the translation types are put in different clusters, meaning that the differences between every pair of translation types are significant.
The ordering of the translation types corroborates that seen in the pairwise analysis (see Section~\ref{s:pairwise_rankings}), namely human translations come on top, followed by NMT outputs and finally, in third place, PBSMT outputs.

If we consider PBSMT's score as a baseline, the score given to the human translations as a gold standard, and the distance between the two as the potential room for improvement for MT, 
we could interpret NMT's score as the progress made in our journey towards better translation quality for novels, departing from PBSMT and targeting human translations as the goal to be reached ultimately.
Using this analogy, although there is still a long way to go, with NMT we have covered already a considerable part of the journey: 
20\%, 22\% and 18\% for Orwell's, Rowling's and Salinger's books, respectively.


%
%

\section{Conclusions and Future Work}\label{s:conclusions}

This chapter has assessed the quality attainable for novels by the two most common paradigms to MT at present, NMT and PBSMT.
To this end, we  built the first in-domain PBSMT and NMT systems for literary text by training them on large amounts of parallel novels.
We  then  automatically evaluated the translation quality of the resulting systems on a set of 12 widely known novels spanning from the 1920s to the present day.
The results  proved favourable for NMT, which outperformed PBSMT by a significant margin for all the 12 novels.

We then delved deeper into the results by analysing the effect of three features of each novel: its lexical richness, its degree of novelty with respect to the training data, and its average sentence length.
Only for the last feature did we find a meaningful correlation against NMT relative improvement over PBSMT, which corroborates the tendency for  improvements in NMT over PBSMT to decrease with sentence length.
This seems to be the main reason behind NMT achieving a relatively low improvement over PBSMT for one of the novels, but we note that particular novel to be an outlier in terms of sentence length.

We have also conducted a human evaluation, where we manually ranked the translations produced by NMT, PBSMT, as well as the human translations for three of the books.
Again, NMT outperformed PBSMT.
For two out of the three books native speakers perceived NMT translations to be of equivalent quality to those of human translations in around one third of the cases (one sixth for PBSMT).

As for future work, we would like to assess the feasibility of using MT to assist with the translation of literary text.
To that end, we plan to carry out an experiment in which we integrate MT into the workflow of professional literary translators by means of post-editing and assess its impact in the translation process (e.g. temporal and technical effort) as well as in the translation result (e.g. quality and reading experience of the resulting translation).

\bibliographystyle{spbasic}
\bibliography{mt_quality_literary_text}

\begin{acknowledgement}
Carme Armentano and Álvaro Bellón ranked the translations used for the human evaluation.
The  research  leading  to  these  results  has  received funding from the European Association for Machine Translation through its 2015 sponsorship of activities programme (project PiPeNovel). The second author is supported by the ADAPT Centre for Digital Content Technology, funded under
the SFI Research Centres Programme (Grant 13/RC/2106).
We would like to thank the Center for Information Technology of the University of Groningen and the Irish Centre for High-End Computing (\url{http://www.ichec.ie}) for providing  computational  infrastructure.
\end{acknowledgement}

\end{document}